\documentclass{article}
\usepackage[margin=2.8cm]{geometry}
\usepackage{graphicx}
\usepackage{cite}
\usepackage{picinpar}
\usepackage{amsmath}
\usepackage{url}
\usepackage{flushend}
\usepackage[latin1]{inputenc}
\usepackage{colortbl}
\usepackage{soul}
\usepackage{booktabs}
\usepackage{multirow}
\usepackage{pifont}
\usepackage{color}
\usepackage{alltt}
\usepackage[hidelinks]{hyperref}
\usepackage{enumerate}
\usepackage{siunitx}
\usepackage{breakurl}
\usepackage{epstopdf}
\usepackage{pbox}
\usepackage{algorithm}
\usepackage{algpseudocode}
\usepackage{amssymb}
\usepackage{subcaption}
\usepackage{palatino}

\usepackage{fancyhdr}
\fancypagestyle{firstpage}{\fancyhead[L]{To appear in Neural Computing and Applications}}

\newcommand{\furl}[1]{\footnote{\url{http://#1}}}

\providecommand{\keywords}[1]
{
  \small	
  \textbf{\textit{Keywords---}} #1
}

\begin{document}

\title{Suicidal Ideation and Mental Disorder Detection with Attentive Relation Networks}

\author{Shaoxiong Ji$^{\S}$\thanks{The work was done while this author was at the University of Queensland.}~,
	Xue Li$^{\dag}$, 
	Zi Huang$^{\dag}$, 
	Erik Cambria$^{\ddag}$ \\
	\small $^{\S}${Department of Computer Science, Aalto University, Finland} \\ \small Email:~shaoxiong.ji@aalto.fi \\
	\small $^{\dag}${School of Information Technology and Electrical Engineering, The University of Queensland, Australia} \\ \small Email:~\{xueli;~huang\}@itee.uq.edu.au \\
	\small $^{\ddag}${School of Computer Science and Engineering, Nanyang Technological University, Singapore} \\ \small Email:~cambria@ntu.edu.sg
	}

\date{}

\maketitle
\thispagestyle{firstpage}

\begin{abstract}
Mental health is a critical issue in modern society, and mental disorders could sometimes turn to suicidal ideation without effective treatment. Early detection of mental disorders and suicidal ideation from social content provides a potential way for effective social intervention. However, classifying suicidal ideation and other mental disorders is challenging as they share similar patterns in language usage and sentimental polarity. This paper enhances text representation with lexicon-based sentiment scores and latent topics and proposes using relation networks to detect suicidal ideation and mental disorders with related risk indicators. The relation module is further equipped with the attention mechanism to prioritize more critical relational features. Through experiments on three real-world datasets, our model outperforms most of its counterparts.
\end{abstract}

\section{Introduction}
Mental health is a global issue, especially severe in most developed countries and many emerging markets. 
According to a report on mental health from the World Health Organization\footnote{Mental health action plan 2013 - 2020, available at \url{http://www.who.int/mental_health/action_plan_2013/mhap_brochure.pdf}}, 1 in 4 people worldwide suffer from mental disorders to some extent. 
Furthermore, 3 out of 4 people with severe mental disorders do not receive treatment, making the problem worse. 
Previous studies reveal that suicide risk usually has a connection to mental disorders~\cite{windfuhr2011suicide}.
Due to severe mental disorders, 900,000 persons commit suicide each year worldwide, making suicide the second most common cause of death among the young. 
Suicide attempters are also reported as suffering from mental disorders, with an investigation on the shift from mental health to suicidal ideation conducted by language and interactional measures~\cite{de2016discovering}.
The US National Alliance on Mental Illness reported that 46\% of suicide victims had experienced mental health conditions\footnote{USA National Alliance on Mental Illness (NAMI) report on Risk Of Suicide, available at \url{https://www.nami.org/Learn-More/Mental-Health-Conditions/Related-Conditions/Suicide}}.
According to the World Bank, at least 10 percent of the global population suffers from mental health issues\footnote{Records updated on Apr 02, 2020. Available at \url{https://www.worldbank.org/en/topic/mental-health}}.

With the advance of social network services, people begin to express their feelings in the forums and seek online support. Traditional ways of prevention include conversation-based consultation and psychological intervention.
However, due to the scarcity and inequality of public resources in health services~\cite{jacob2014classification}, many victims could not get effective treatments even though some of them are suffering from severe mental disorders. 

Transferring from mental disorders to suicidal ideation and the final suicide action is a long-term process. Gilat et al.~\cite{gilat2011offering} scaled suicide risks into four levels, i.e., non-suicidal, suicidal thoughts or wishes, suicidal intentions, and suicidal act or plan.
Before suicidal ideation, victims may suffer from different kinds of other mental disorders. According to meta-analyses, underlying mental disorders can lead to suicide, mainly in high-income countries with a figure of 90\%.
According to meta-analyses by Hannah Ritchie and Max Roser\footnote{Published online at OurWorldInData.org. Available at \url{https://ourworldindata.org/mental-health}}, underlying mental disorders can lead to suicide. This is especially prevalent in high-income countries, with 90\% of people who encounter mental health issues have suicidal thoughts.
The social networking service has become one of the most valuable tools to provide support and feedback for people with mental health issues~\cite{shepherd2015using}.
To provide effective early suicide prevention given limited support resources, it is necessary to triage the risk levels automatically and provide conversational support accordingly to relieve victims' issues~\cite{ji2021suicidal}.
Our motivation is to use deep learning techniques to enable early detection and identify people's risk levels, which can help the social workers or experts to have a prior understanding of people's situation when trying to relieve their mental health issues. 
The automatic detection technique can be applied to mental health monitoring and help to facilitate online support.
We conduct suicidal ideation and mental health detection, aiming to distinguish early-stage mental disorders and severe suicidality automatically. This classification of suicide risk and mental status could help social workers prioritize and allocate resources to people with different needs and situations according to their severity. Thus, effective prevention measures can be taken to stop mental health discourse transitions to suicidal thoughts.

Suicide and mental health issues could be categorized as different levels, taken as a multi-class classification problem. There are many types of mental disorders according to two main diagnostic schemes for identifying mental disorders, i.e., Diagnostic and Statistical Manual of Mental Disorders (DSM-5)\furl{psychiatry.org/psychiatrists/practice/dsm} and Chapter V Mental and Behavioral Disorders of International Statistical Classification of Diseases and Related Health Problems 10th Revision (ICD-10)\furl{apps.who.int/classifications/icd10/browse/2016/en\#/V}.
Recent works using deep neural networks have revolutionized this field of text classification.
However, classifying mental health and suicidal ideation is a more specific task that requires focusing on potential victims' language usage.
Suicidal ideation and mental disorders (e.g., depression, anxiety, and bipolar) in online social content share quite similar patterns, including the language usage~\cite{coppersmith2015quantifying}, topic distribution~\cite{ji2018supervised}, and sentimental polarity~\cite{chen2018mood} . 
Sentiment analysis is studied for assigning emotions to suicide notes~\cite{mccart2012using}, where most of the notes contain many negative expressions. 
Topics including job stress, family issues, and personal crisis are pretty common among those posts~\cite{ji2018supervised}. 
Thus, classifying suicidal ideation and other mental health issues requires attention to understand the subtle differences among those characteristics. 
Noticing that people's posting showing feelings or expressing suffering contains their sentiment to some extent, we propose to capture this valuable and vital information to learn richer sentence representation and better encode risk actions and people's mental or social state.

This paper investigates deep learning-based models for text classification on some existing and self-collected datasets from social networking websites. It proposes an enhanced relation network (RN) to provide a more accurate classification of suicide risk levels and suicidal ideation vs. other mental disorders such as depression and anxiety.
Relation networks~\cite{raposo2017discovering} are firstly proposed for visual reasoning, while the undefined relationship between texts and sentiment lexicons or relationship between texts and topic distributions remains unexplored. In this paper, we firstly migrate the principle of visual relational reasoning to relate extracted informative features and hidden text representation and develop a novel attentive encoding model to capture the relation between suicide indicators such as sentiments and event topics and text mentions. This novel relational encoding model can reason over the risk indicators and sentence embeddings and learn richer representations.

Our contributions could be summarized as:
\begin{itemize}
\item This paper focuses on identifying suicidal ideation and different kinds of mental disorders for early warning. Specifically, we consider both user-level and post-level detection. 
\item To improve risk identification performance, we propose an attentive RN model with text representation and two sets of risk indicators encoded, i.e., lexicon-based sentimental state and latent topics within posts.
\item Experiments on public datasets and our collected dataset show that our proposed method can improve the predictive performance. 
\end{itemize}

This paper is organized as follows. Related work on mental disorders, suicidal ideation, text classification, and relational reasoning are reviewed in Section~\ref{sec:related}. In Section~\ref{sec:methods}, we introduce the proposed method that introduces sentimental lexicon and topic model into relational encoding with attentive RN. Datasets are introduced in Section~\ref{sec:data}, together with a simple exploratory analysis. Experimental settings and results are presented in Section~\ref{sec:exp}.
In Section~\ref{sec:conclusion}, we conclude and have a brief outlook for future work. 

\section{Related~Work}
\label{sec:related}
This paper focuses on enabling effective relational text encoding to classify suicidal ideation and mental disorders. Related works include research on mental disorders, suicidal ideation, text classification techniques, and relational reasoning.

Mental health issues and suicidal ideation~\cite{ji2021suicidal} have been studied including the clinical interaction~\cite{venek2017adolescent}, classifying self-report screening questionnaire~\cite{gomez2012suicide}, and detection from the data mining perspective~\cite{benton2017multi, ji2019detecting}. With the popular social text analysis and natural language processing techniques, more and more research turns to investigate the mental health discourse~\cite{pavalanathan2015identity}, discover self-harm content~\cite{wang2017understanding}, and detect social network mental disorders~\cite{shuai2016mining}, depression~\cite{de2013predicting} and suicidal ideation~\cite{ji2018supervised} in social media. 
Li et al.~\cite{li2018text} detected the changes of online users in the mental health communities of social media. 
Cao et al.~\cite{cao2020building} proposed to use a personal knowledge graph to improve the detection performance.
Affective information is widely used for mental health and suicidality detection. Nguyen et al.~\cite{nguyen2014affective} proposed a thorough affective analysis and a content analysis between depression and control communities. Ren et al.~\cite{ren2016examining} proposed an accumulated emotion model to classify suicide blog stream. Chen et al.~\cite{chen2018mood} measured emotions to identify depression on Twitter. 
We recommend readers to check out the recent review article~\cite{ji2021suicidal} about suicidal ideation detection with a detailed introduction and summary.

Suicidal ideation and mental disorder detection are technically formulated as a text classification, which has experienced rapid development with the development of deep neural networks~\cite{yourec}. 
Kim~\cite{kim2014convolutional} proposed convolutional neural networks for sentence classification. To capture long term dependencies in sentences, the long short-term memory (LSTM)~\cite{hochreiter1997long} was applied. Lai et al.~\cite{lai2015recurrent} proposed recurrent convolutional neural networks combining two popular neural network architectures for text classification. Li et al.~\cite{liigen} combined reinforcement learning, generative adversarial networks, and recurrent neural networks for text categorization. Zhao et al.~\cite{zhatow} explored the use of capsule networks for challenging natural language processing applications. The attention mechanism~\cite{bahdanau2014neural} is also widely used in text classification. For example, Lin et al.~\cite{lin2017structured} proposed self-attention to learn structured sentence embedding, and Ma et al.~\cite{maatar} proposed an attentive LSTM for aspect-based sentiment analysis.

Those informative affective cues and neural advances inspire this paper to develop a deep model to learn hidden text representation and encode extracted features such as emotion information~\cite{chen2018mood} and topic descriptions~\cite{ji2018supervised}, and external resources like domain-specific lexicons~\cite{hamilton2016inducing} in a hybrid manner. 
This paper incorporates relational reasoning with relation networks (RNs) to inject and fuse that multi-channel auxiliary information and rich textual representations. RNs are initially utilized for scene object discovery by exploiting relations among objects~\cite{raposo2017discovering}, and further introduced to relational reasoning for visual question answering by calculating the relation score of the feature maps of object pairs and question representation~\cite{santoro2017simple}. 
As for our application scenario of suicidal ideation detection, it is critical to understand the relation between suicidality and risk indicators such as individuals' sentiment and life events. This is also the novelty of this paper, i.e., to reason over the relation between risk indicators between text mentions using relation networks.

\section{Methods} 
\label{sec:methods}

\subsection{Problem~Definition}
Detecting suicidal ideation and mental disorders in social content is technically a domain-specific task of text classification. Our paper conducts a fine-grained suicide risk assessment and classification of multiple mental health issues, which are naturally regarded as multi-class classification. For fine-grained suicide risk, the risk levels include none, low, moderate, and severe risk, while for mental health classification, specific mental disorders are depression, anxiety, and bipolar. Moreover, there are two subtasks for specific social content settings, i.e., post-level and user-level classification. The former one takes single post $p$ as input, while the latter one detects the suicide attempter with multiple posts $P=\{p_1, p_2, \dots, p_n\}$.

\subsection{Model~Architecture}
\begin{figure}[ht!]
\begin{center}
\includegraphics[width=0.4\textwidth]{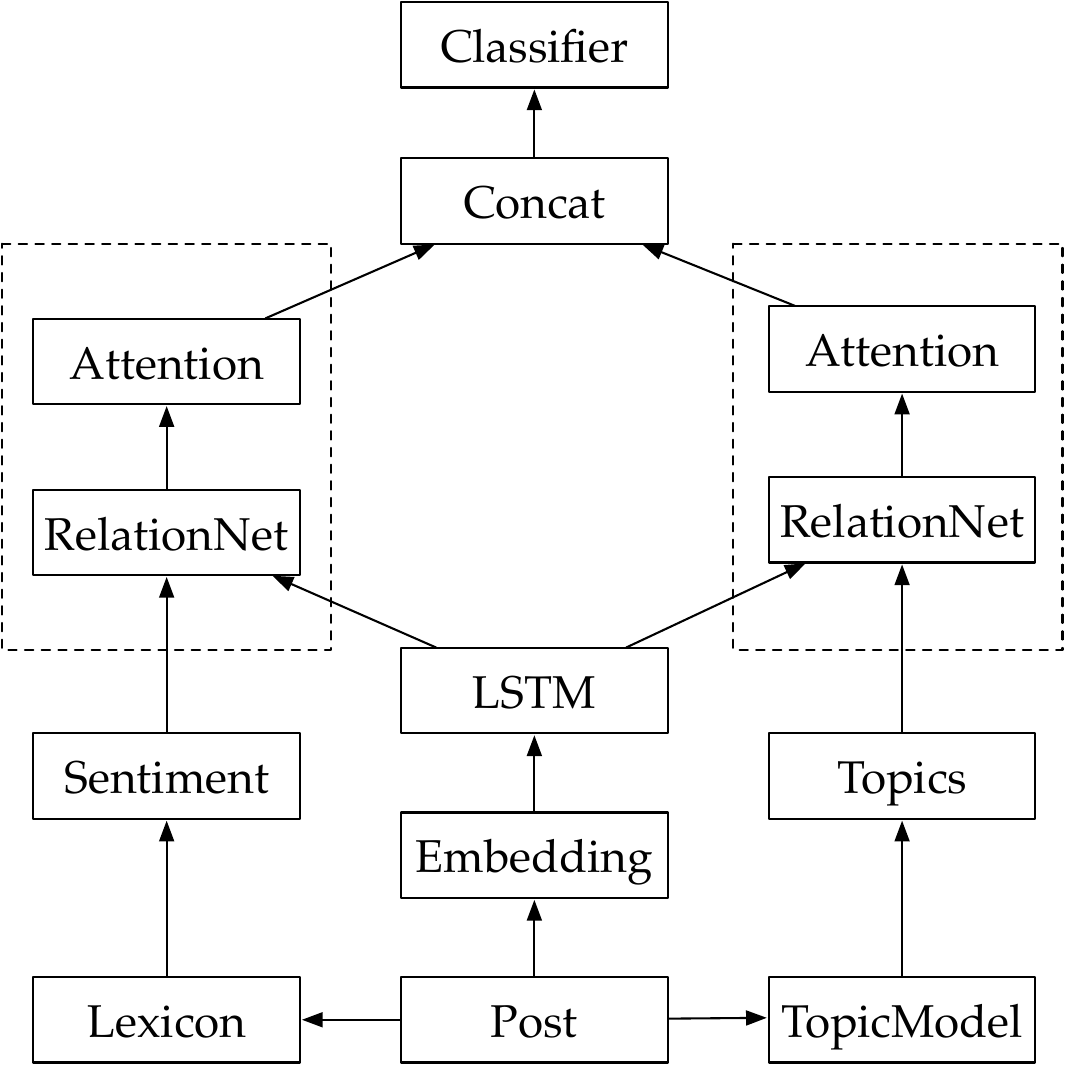}
\caption{The architecture of the proposed model}
\label{fig:architecture}
\end{center}
\end{figure}

The proposed model consists of two steps, i.e., post representation and relational encoding module as illustrated in Fig.~\ref{fig:architecture}. The post representation includes two parts of extraction of risk-related state indicators and LSTM text encoder. The relation module, as shown in the dashed box of Fig.~\ref{fig:architecture} utilizes a vanilla RN for reasoning on the connection between state indicators and user's posts and the attention mechanism for prioritizing more important relation scores of text encoding.

\subsection{Text~Encoding~and~Risk~Indicators}
User post sequence is embedded into word vectors of $p=\{w_1, w_2, \dots, w_n\} \in \mathbb{R}^{l \times d}$, where $l$ is the length of posts and $d$ is the dimension of word embeddings. 
We apply bidirectional LSTM in Eq.~\ref{eq:bilstm} for text encoding to capture the adjacent dependency of words. 
\begin{equation}
\label{eq:bilstm}
\begin{array}{l}
{\overrightarrow{h_{t}}=\overrightarrow{\operatorname{LSTMcell}}\left(w_{t}, \overrightarrow{h_{t-1}}\right)} \\ 
{\overleftarrow{h_{t}}=\overleftarrow{\operatorname{LSTMcell}}\left(w_{t}, \overleftarrow{h_{t+1}}\right)}
\end{array}
\end{equation}
The hidden state is obtained by concatenating each direction as $h_t=\operatorname{concat}(\overrightarrow{h_{t}}, \overleftarrow{h_{t}})$, where $h_t\in \mathbb{R}^{l\times 2n}$ given $n$ as the number of hidden units. 

Sentimental information plays an important role when people are expressing their sufferings and feelings in online social networks. 
To measure the sentiment, we take sentiment lexicons as additional information. Specifically, domain-specific sentiment lexicons~\cite{hamilton2016inducing} from communities in Reddit is used. Seed words induce the sentiment lexicons with domain-specific word embedding and a label propagation framework. For the details of building domain-specific sentiment lexicons,  we recommend that readers refer to the original paper. 
The extracted sentiment information of post denoted as $s\in \mathbb{R}^{l}$ acts as a state indicator representing post creators' internal sentimental state. 
Correspondingly, extrinsic indicators such as people's event topics reveal another dimension as the risk indicator. We introduce a topic model to learn unsupervised topical features to capture external factors of suicidal ideation or mental disorders. Specifically, Latent Dirichlet Allocation (LDA)~\cite{blei2003latent} is applied to extract latent topics in social posts to represent people's sufferings such as life events, social exposure, and other experience in the real world. The probability score vectors of posts belonging to all extracted topics are represented as $v \in \mathbb{R}^m $, where $m$ is the number of topics. 

\subsection{Relation Network with Attention}
RN~\cite{santoro2017simple} is a neural module for relational reasoning. It is originally proposed to capture the relation between objects. Given objects of $\mathcal{O}=\left\{o_{1}, o_{2}, \dots, o_{n}\right\}$ and functions of $f_\phi$ and $g_\theta$, a RN is defined in Eq.~\ref{eq:relation}. The output of $g_\theta$ is called the learned ``relation'', while the $f_\phi$ function acts as the classifier. 
\begin{equation}
\label{eq:relation}
\mathrm{RN}(\mathcal{O})=f_{\phi}\left(\sum_{i, j} g_{\theta}\left(o_{i}, o_{j}\right)\right)
\end{equation}

\begin{figure*}[htbp]
\begin{center}
\includegraphics[width=0.9\textwidth]{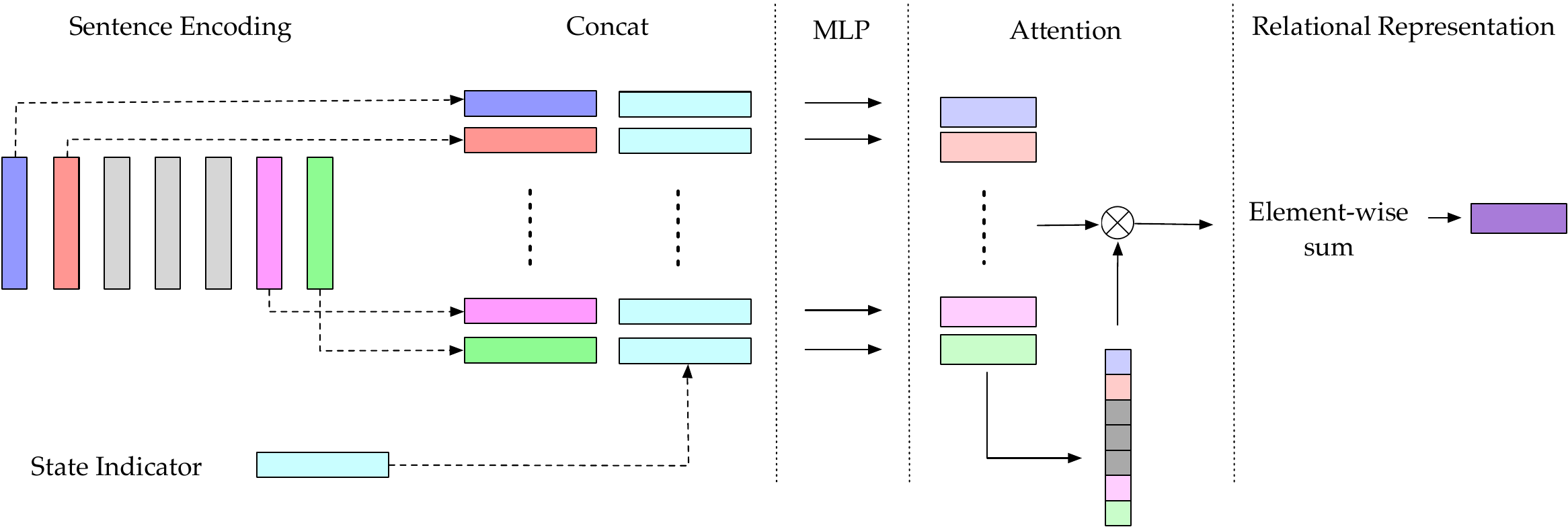}
\caption{Relation network with attention mechanism}
\label{fig:relation}
\end{center}
\end{figure*}

We aim to encode risk factors of suicidal ideation and mental disorders into textual representation. Thus, we take text encoding and state indicators as the input of RNs to calculate relation scores between each token in posts and state indicators modeled by sentiment and topic features. The attention mechanism is further incorporated with the relation module by assigning attention weights to the learned relations. The idea of attentive RN is shown in Fig.~\ref{fig:relation}. 
The text representation is encoded by an LSTM network, which captures the sequential independence. The encoded text representation is then concatenated with the representations of state indicators. Here, we consider two indicators of sentiment and topic features, with the expanded representations by repeating the extracted vectors denoted as $S=[s, s, \dots, s] \in \mathbb{R}^{l\times l}$ and $V=[v, v, \dots, v] \in \mathbb{R}^{l\times m}$ respectively. Then, they are inputted into RNs to calculate relation vector $r_i \in\mathbb{R}^{k}$, where $k$ is the dimension of hidden representation, with a multiple layer perception (MLP) as in Eq.~\ref{eq:MLP} for the sentiment indicator.
\begin{equation}
\label{eq:MLP}
r_i = \operatorname{MLP}(h_i, s_i) 
\end{equation}
The attention is calculated as follows:
\begin{equation}
\alpha=\operatorname{softmax}\left( \left[r_{1}, r_{2}, \ldots, r_{l}\right]W^T + b \right),
\end{equation}
where $W\in\mathbb{R}^{1\times k}$, $b\in \mathbb{R}^l$ and $\alpha \in \mathbb{R}^l$. 
By element-wise product, the attentive representation of learnt relations can be calculated as
\begin{equation}
\tilde{r} =\alpha \otimes \left[r_{1}, r_{2}, \ldots, r_{i}\right]
\end{equation}
where $\tilde{r}\in \mathbb{R}^{l\times k}$. By applying element-wise sum over $\tilde{r}$, we get the final attentive relational representation.

\subsection{Classification}
The last step is to use the learned representation, which contains sequential information and risk indicators for classification. Specifically, we concatenate relational representations of $e=[\tilde{r}_s, \tilde{r}_v]$ from two channels as shown in Fig.~\ref{fig:architecture}, and use the fully connected layer with non-linear activation function of $f(\cdot)$ to produce the logits for prediction as follows. 
\begin{equation}
\begin{array}{rlr}
{l} & {=\operatorname{f}\left(W_l e+b_l\right)} \\ 
{\mathcal{P}} & {=\operatorname{softmax}\left(W_o l+b_{o}\right)} \\ 
\end{array}
\end{equation}
where $W_{l} \in \mathbb{R}^{d_{l} \times d_{e}}, b_{l} \in \mathbb{R}^{d_{l}}, W_{o} \in \mathbb{R}^{c \times d_{l}}, b_{o} \in \mathbb{R}^{c}, \mathcal{P} \in \mathbb{R}^{c}$
For multi-class classification, the predicted label is produced by 
\begin{equation}
\hat{y} =\underset{i}{\operatorname{argmax}} \left(\mathcal{P}_i\right)
\end{equation}

\subsection{Training}
Our proposed model has two training phases, i.e., training the LDA topic model and the classification model. 
For the topic model, LDA assumes a generative process of documents as random mixtures over latent topics. A topic can be inferred as a distribution over the words, where the Bayesian inference is used for learning various distributions. In practice, we use the Gensim library\furl{radimrehurek.com/gensim} to build the topic model during implementation.

For the ultimate target of suicide ideation and mental health detection, we use the cross-entropy loss with L2 regularization, denoted as: 
\begin{equation}
L=-\frac{1}{\sum_{s=1}^{N} c(s)} \sum_{i=1}^{N} \sum_{j=1}^{c(i)} \log \mathcal{P}_{i, j}\left[y_{i, j}\right]+\lambda\|\theta\|_{2},
\end{equation}
where $c(s)$ is the set of labels, $\theta$ represents all the trainable parameters, and $\lambda$ is the regularization coefficient or the so-called weight decay rate. 
We apply the Adam algorithm~\cite{kingma2014adam} to optimize the objective function.

\section{Data}
\label{sec:data}
We use data from two popular social networking websites, i.e., Reddit and Twitter, with three datasets derived. Two of them are from Reddit, with one public dataset and one firstly collected in this paper. People's posts from an active subreddit for online support in Reddit, called ``SuicideWatch''(SW)\furl{reddit.com/r/SuicideWatch}, are intensively used in these two datasets. The last one is collected from Twitter by combining several existing data sources. These datasets cover suicide and other mental health issues, with specific categories reported in ICD-10 as listed in Table~\ref{tab:definition}.

\begin{table}[htp]
\scriptsize
\caption{Description of mental disorders in ICD-10}
\begin{center}
\begin{tabular}{|c|p{6.3cm}|}
\toprule
Category & Descriptions mentioned in ICD-10 \\
\midrule
Suicide & Intentional self-harm, suicidal ideation (tendencies) \\
Depression & In typical mild, moderate, or severe depressive episodes, the patient suffers from lowering of mood, reduction of energy, and decrease in activity. \\
Anxiety & Phobic anxiety and other anxiety disorders \\
Bipolar & A disorder characterized by two or more episodes in which the patient's mood and activity levels are significantly disturbed\\
PTSD & Arises as a delayed or protracted response to a stressful event or situation (of either brief or long duration) of an exceptionally threatening or catastrophic nature\\
\bottomrule
\end{tabular}
\end{center}
\label{tab:definition}
\end{table}%

\subsection{UMD Reddit Suicidality Dataset}
The UMD Reddit Suicidality Dataset~\cite{shing2018expert} was collect from anonymous discussion forums in \url{Reddit.com}. It contains posts of 620 users in the training set and 245 users in the testing set sampled from 11,128 users in the subreddit ``SuicideWatch'' and 11,129 users in other subreddits. 
It is annotated by crowdsourcing workers and human experts via a crowdsourced platform, referring to the original paper for annotations' details. The suicide risk is scaled to four levels, i.e., no risk(a), low(b), moderate(c), and severe risk(d). It also provides coarse labels where no risk and low risk are given the label of 0, moderated, and severe risk are labeled as 1, together with the control group as the label of -1.

This dataset was released as the CLPsych 2018 Shared Task~\cite{shing2018expert}, and then a new version of it acted as the CLPsych 2019 Shared Task~\cite{zirikly2019clpsych}.
This paper uses a dataset derived from the UMD dataset in user-level four categories of suicide risk. The statistical information of the dataset is illustrated in Tab.~\ref{tab:umd_info}. Besides, we include control users (labeled as ``None'') into this annotation set. 

\begin{table}[!htbp]
\scriptsize
\caption{Statistical information of UMD Reddit Suicidality Dataset}
\label{tab:umd_info}
\centering
\begin{tabular}{| c | c | c |}
\toprule
Annotation & Numbers & \% of a/b/c/d/ levels \\
\midrule
crowd & 621 & 26\%/10\%/24\%/40\% \\
expert & 245 & 29\%/9\%/25\%/37\%\\
\bottomrule
\end{tabular}
\end{table}

We use the transformed labels from the raw label according to the original description of this dataset. Specifically, raw labels of ``c'' or ``d'' are transformed into 1, raw labels of ``a'' or ``b'' are transformed into 0, and the label of a control user is -1 by definition. We split the whole dataset into training, validation, and testing sets as listed in Table~\ref{tab:umd_used}.

\begin{table}[htp]
\scriptsize
\caption{Statistical information of UMD dataset with train/validation/test split}
\begin{center}
\begin{tabular}{|c|c|c|c|}
\toprule
Label & \#/\% of train & \#/\% of valid. & \#/\% of test \\
\midrule
-1 & 495/49.8489\% & 126/50.6024\% & 245/50.000\% \\
0 & 188/31.2185\% & 89/35.7430\% & 86/17.551\% \\
1 & 310/18.9325\% & 34/13.6546\% & 159/32.449\% \\
\bottomrule
\end{tabular}
\end{center}
\label{tab:umd_used}
\end{table}%

\subsection{Reddit~SWMH~Dataset}
As severe mental health issues are very likely to lead to suicidal ideation, we also collect another dataset from some mental health-related subreddits in \url{Reddit.com} to further the study of mental disorders and suicidal ideation. We name this dataset as Reddit SuicideWatch and Mental Health Collection, or SWMH for short, where discussions comprise suicide-related intention and mental disorders like depression, anxiety, and bipolar. We use the Reddit official API\furl{reddit.com/dev/api} and develop a web spider to collect the targeted forums. This collection contains a total of 54,412 posts. 
Specific subreddits are listed in Table~\ref{tab:reddit_info}, as well as the number and the percentage of posts collected in the train-val-test split. 

In those communities or so-called subreddits, people discussed their own or their relative's mental disorders and sought advice or help. We perform experimental analysis on this dataset to identify discussions about suicidality and mental disorders.

\begin{table}[htp]
\scriptsize
\caption{Statistical information of SuicideWatch and mental health related subreddits, i.e., SWMH dataset}
\begin{center}
\begin{tabular}{|c|c|c|c|}
\toprule
Subreddit & \#/\% of train & \#/\% of valid. & \#/\% of test \\
\midrule
depression 	& 11,940/34.29 		& 3,032/34.83 	& 3,774/34.68   	\\
SuicideWatch 	& 6,550/18.81 		& 1,614/18.54 	& 2,018/18.54 		\\
Anxiety    	& 6,136/17.62 		& 1,508/17.32 	& 1,911/17.56 		\\
offmychest   	& 5,265/15.12 		& 1,332/15.30 	& 1,687/15.50		\\
bipolar     	& 4,932/14.16 		& 1,220/14.01 	& 1,493/13.72		\\
\bottomrule
\end{tabular}
\end{center}
\label{tab:reddit_info}
\end{table}%

\subsection{Twitter~Datasets~Collection}
The third dataset is a collection of different subsets from Twitter with an overlapping check. Sampled instances from two datasets consist of most samples of this dataset. First, 594 instances of tweets containing suicidal ideation are from Ji et al.~\cite{ji2018supervised}, with additional 606 tweets manually collected by this work. Second, the same number of depression and post-traumatic stress disorder (PTSD) posts are sampled from CLPsych 2015 shared task dataset~\cite{coppersmith2015clpsych}. This dataset is available upon request\footnote{Request for data access via \url{http://www.cs.jhu.edu/~mdredze/datasets/clpsych_shared_task_2015/}.}. 
Last, the control group where Twitter users are not identified as having a mental condition or suicidal ideation is comprised by sampling regular tweets from previously mentioned datasets~\cite{ji2018supervised, coppersmith2015clpsych}. Finally, this Twitter dataset collection contains 4,800 tweets with four classes of suicidality, depression, PTSD, and control. 

\subsection{Linguistic Clues and Emotion Polarity}
We have a brief exploratory analysis of the data. Some selected linguistic, statistical information of UMD dataset extracted by Linguistic Inquiry and Word Count software (LIWC)\furl{liwc.wpengine.com} is shown in Table ~\ref{tab:liwc}. The risk of suicide increases among labels of -1, 0, and 1. The linguistic inquiry results show that negative emotion, anxiety, and sadness are expressed more in posts with high-level suicide risk. The same trend exists in family issues, death-related mentions, and swear words. Naturally, positive emotions are less presented in posts with high suicide risk.

\begin{table}[!ht]
\scriptsize
\caption{Selected linguistic statistical information of UMD dataset extracted by LIWC}
\label{tab:liwc}
\centering
\begin{tabular}{|l|r|r|r|}
\toprule
Linguistic clues	&	Label -1	&	Label 0	&	Label 1	\\
\midrule
positive emotion	&	3.30	&	3.12	&	2.96	\\
negative emotion	&	1.56	&	2.30	&	2.74	\\
anxiety	&	0.17	&	0.33	&	0.41	\\
sadness	&	0.28	&	0.50	&	0.68	\\
\hline
family	&	0.29	&	0.39	&	0.47	\\
friend	&	0.43	&	0.56	&	0.54	\\
work		&	2.54	&	1.92	&	1.80	\\
money	&	1.13	&	0.71	&	0.61	\\
death	&	0.22	&	0.29	&	0.36	\\
\hline
swear words	&	0.23	&	0.33	&	0.40	\\
\bottomrule
\end{tabular}
\end{table}

\section{Experiments}
\label{sec:exp}
We compare it with several text classification models on three real-world datasets to evaluate our proposed model's performance. Baselines and empirical settings are introduced, and results are reported and discussed in this section. 
\subsection{Baseline~and~Settings}

We compared five popular classification models with our proposed method. These baseline models are described as follow:
\begin{itemize}
\item \textbf{fastText}~\cite{joulin2016bag}: an efficient text classification model with bag of words sentence representation and a linear classifier. 
\item \textbf{CNN}~\cite{kim2014convolutional}: it applies convolutional neural networks over the word embedding of a sentence to produce feature maps and then uses max-pooling over the features.
\item \textbf{LSTM}~\cite{hochreiter1997long}: it takes sequential word vectors as input to the recurrent LSTM cells applies pooling over the output to obtain final representation. By combining the forward and backward direction, it becomes bidirectional LSTM (BiLSTM). 
\item \textbf{RCNN}~\cite{lai2015recurrent}: this model at first applies LSTM model~\cite{hochreiter1997long} to capture sequential information, and then applies CNN~\cite{kim2014convolutional} to further extract features. It has a bidirectional version using BiLSTM. 
\item \textbf{SSA}~\cite{lin2017structured}: it proposed a structured self-attention mechanism with multiple hops by introducing a 2D matrix for embedding representation. The self-attention is applied to the sequential hidden states of the LSTM network. 
\end{itemize}

All the baseline models and our proposed method are implemented by PyTorch\furl{pytorch.org} and run in a single GPU (Nvidia GeForce GTX 1080 Ti). We train the models for 50 epochs by default, setting the batch size 128 and 16 according to the size of datasets. Specifically, the UMD dataset's batch size is 16, and for SWMH and Twitter data collection, the batch size is 128. 
We use pretrained GloVe~\cite{pennington2014glove} word representation with either static or dynamic embedding utilized for the word embedding. Our proposed method enumerates all the 250 subreddit lexicons of Reddit and the number of topics from 5 to 20. 
We select the best validation performance in multiple trials and report the testing performance as experimental results.

Our RN-based model uses an additional attentive relation network, and SSA applies a structured attention network. 
Thus, SSA and RN have more trainable parameters compared with vanilla CNN and LSTM. 
Our model size is on par with the SSA model. 
In our experimental setting, we use 100D word embeddings and set the hidden dimension of the LSTM unit to 300. 
As our model uses another MLP to calculate the relation score, it consumes more parameters, around 30K parameters, than the SSA model. 
However, it is worth sacrificing the model size to achieve higher predictive performance for an essential mission of saving lives.

Automatic detection aims to produce effective diagnoses (i.e., true positive) and decrease the incorrect diagnoses (i.e., false positive) to avoid patients' stress and anxiety caused by false detection. Thus, during the evaluation process, we report both prediction accuracy and the weighted average F-score metric. For unbalanced datasets, we apply weight penalty to the objective function and report the weighted average results.

\subsection{Results}
\label{sub:res}
We evaluate the experimental performance on three datasets collected from Reddit and Twitter. For the UMD Suicidality dataset and the SWMH dataset, the reported results are weighted average. 

\subsubsection{UMD~Suicidality}
We firstly implement our method and baselines on the UMD suicidality dataset for user-level classification. When processing a set of posts from users, all posts of users are concatenated as user-level representation.  Concatenated posts are truncated or padded with zeros to ensure an identical dimensionality. The results of four metrics, including accuracy, precision, recall, and F1-score, are very close for RN and BiLSTM. The BiLSTM model gains the highest accuracy of 56.94\%, and our model follows at the second place of 56.73\%. Our model equipped with relational encoding, however, has a higher F1 score than all the baselines. Noticing these very close results, we then further analyze each class's results in Section~\ref{: each-class}. 

\begin{table}[!htbp]
\scriptsize
\caption{Comparison of different models on UMD dataset for user-level classification, where precision, recall, and F1 score are weighted average.}
\label{tab:umd}
\centering
\begin{tabular}{| c | c | c | c | c |}
\toprule
Model	&	Accuracy	&	Precision	&	Recall	&	F1	\\
\midrule
fastText	&	0.5327	&	0.5300	&	0.5327	&	0.5202	\\	
CNN	&	0.5531	&	0.4498	&	0.5531	&	0.4935	\\	
LSTM	&	0.5612	&	0.4625	&	0.5612	&	0.5071	\\	
BiLSTM	&	\textbf{0.5694}	&	0.5029	&	\textbf{0.5694}	&	0.5233	\\	
RCNN	&	0.5592	&	0.4953	&	0.5592	&	0.5111	\\	
SSA	&	0.5633	&	0.4711	&	0.5633	&	0.4839	\\	
RN	&	0.5673	&	\textbf{0.5405}	&	0.5673	&	\textbf{0.5453}	\\	
\bottomrule
\end{tabular}
\end{table}

\subsubsection{Reddit~SWMH}
\label{sec:swmh}
Then, we perform experiments on the Reddit SWMH dataset, which contains both suicidal ideation and mental health issues, to study our model's predictive performance. It is a larger dataset with more instances when compared with the UMD dataset. Experiments on this dataset show RN's relational encoding capacity for mental health-related texts with similar characteristics. As shown in Table~\ref{tab:swmh}, our model beats all baseline models in terms of all the four metrics. 

\begin{table}[!htbp]
\scriptsize
\caption{Comparison of different models on Reddit SWMH collection, where precision, recall, and F1 score are weighted average.}
\label{tab:swmh}
\centering
\begin{tabular}{| c | c | c | c | c |}
\toprule
Model	&	Accuracy	&	Precision	&	Recall	&	F1	\\
\midrule
fastText	&	0.5722	&	0.5760	&	0.5722	&	0.5721	\\
CNN	&	0.5657	&	0.5925	&	0.5657	&	0.5556	\\
LSTM	&	0.5934	&	0.6032	&	0.5934	&	0.5917	\\
BiLSTM	&	0.6196	&	0.6204	&	0.6196	&	0.6190	\\
RCNN	&	0.6096	&	0.6161	&	0.6096	&	0.6063	\\
SSA	&	0.6214	&	0.6249	&	0.6214	&	0.6226	\\
RN	&	\textbf{0.6474}	&	\textbf{0.6510}	&	\textbf{0.6474}	&	\textbf{0.6478}	\\
\bottomrule
\end{tabular}
\end{table}

\subsubsection{Twitter~Collection}
Lastly, we conduct experiments on the Twitter dataset with similar settings to previous experiments. Unlike posts in Reddit, tweets in this dataset are short sequences due to the tweet's length limit of 280 characters. The results of all baseline methods and our proposed method are shown in Table~\ref{tab:twitter}. Among these competitive methods, our model gains the best performance on these four metrics, with 1.77\% and 1.82\% improvement than the second-best BiLSTM model in terms of accuracy and F1-score, respectively. Our proposed method introduces auxiliary information of lexicon-based sentiment and topics learned from the corpus and utilizes RNs to model the interaction between LSTM-based text encodings and risk indicators. Richer auxiliary information and efficient relational encoding help our model boost performance in short tweet classification. 

\begin{table}
\scriptsize
\centering
\caption{Performance comparison on Twitter dataset, where precision, recall, and F1 score are weighted average.}
\label{tab:twitter}
\begin{tabular}{| c | c | c | c | c |}
\toprule
Model	&	Accuracy	&	Precision	&	Recall	&	F1	\\
\midrule
fastText	&	0.7927	&	0.7924	&	0.7927	&	0.7918	\\
CNN		&	0.7885	&	0.7896	&	0.7885	&	0.7887	\\
LSTM	&	0.8021	&	0.8094	&	0.8021	&	0.8039	\\
BiLSTM	&	0.8208	&	0.8207	&	0.8208	&	0.8195	\\
RCNN	&	0.8094	&	0.8089	&	0.8094	&	0.8090	\\
SSA		&	0.8156	&	0.8149	&	0.8156	&	0.8152	\\
RN		&	\textbf{0.8385}	&	\textbf{0.8381}	&	\textbf{0.8385}	&	\textbf{0.8377}	\\
\bottomrule
\end{tabular}
\end{table}

\subsection{Performance on Each Class}
\label{sec:each-class}
This section studies the performance of each class of the UMD dataset. We select two baselines with better performance for comparison as shown in Fig.~\ref{tab:versus}. The proposed RN-based model is flawed in predicting posts without suicidality but good at predicting posts with high suicide risk. Unfortunately, all these three models have an inferior capacity for predicting posts with low suicide risk. In the UMD dataset with a small volume of instances, these models tend to predict posts as classes with more instances, even though we apply penalty on the objective function.

\begin{table}[htp]
\scriptsize
\caption{Performance on each class of UMD suicidality dataset}
\begin{center}
\begin{tabular}{|c|c|c|c|c|}
\toprule
Label	&Metrics	&	BiLSTM	&	SSA	&	RN	\\
\midrule
\multirow{3}{*}{-1}	&Precision	&	0.62	&	0.57	&	0.69	\\
	&Recall	&	0.77	&	\textbf{0.92}	&	0.70	\\
	&F1-score	&	0.69	&	0.70	&	0.69	\\
\hline
	&Precision	&	0.51	&	0.57	&	0.48	\\
1	&Recall	&	0.55	&	0.31	&	\textbf{0.62}	\\
	&F1-score	&	0.53	&	0.41	&	0.54	\\
\hline
	&Precision	&	0.15	&	0.00	&	0.24	\\
0	&Recall	&	0.02	&	0.00	&	0.09	\\
	&F1-score	&	0.04	&	0.00	&	0.13	\\
\bottomrule
\end{tabular}
\end{center}
\label{tab:versus}
\end{table}

\subsection{Ablation~Study}
We then conduct an ablation study to explore several variants and compare their performance. We compare our complete framework with three different settings of injecting risk indicators. The BiLSTM+concat model concatenates the final hidden state with sentiment and topic features. The BiLSTM+RN+sentiment and BiLSTM+RN+topic use single-channel relation network. The results on the Twitter collection are shown in Table~\ref{tab:ablation}. Relation networks are generally better than BiLSTM with simple feature concatenation, where the performance of the latter model decreased compared to vanilla BiLSTM. The BiLSTM+RN+sentiment model is better than vanilla BiLSTM, while the BiLSTM+RN+topic is slightly worse. However, two channels of sentiment and topic couple well, achieving the best performance. This study shows the effectiveness of the proposed model. 

\begin{table}[htp]
\scriptsize
\caption{Performance of different variants of risk indicator injection}
\begin{center}
\begin{tabular}{|c|c|c|c|c|}
\toprule
Model	&	Accuracy	&	Precision	&	Recall	&	F1	\\
\midrule
BiLSTM	&	0.8208	&	0.8207	&	0.8208	&	0.8195 \\															
BiLSTM+concat	&	0.8167	&	0.8262	&	0.8167	&	0.8190	\\
BiLSTM+RN+sentiment	&	0.8240	&	0.8246	&	0.8240	&	0.8239	\\
BiLSTM+RN+topic	&	0.8198	&	0.8177	&	0.8198	&	0.8183	\\
BiLSTM+RN+sent.+topic	&	\textbf{0.8385}	&	\textbf{0.8381}	&	\textbf{0.8385}	&	\textbf{0.8377}	\\
\bottomrule
\end{tabular}
\end{center}
\label{tab:ablation}
\end{table}%

\subsection{Error~Analysis~and~Limitations}
\begin{figure}[!ht]
\caption{Confusion matrix on UMD dataset}
\scriptsize
\centering
\begin{subfigure}[b]{0.44\textwidth}
	\includegraphics[width=\textwidth]{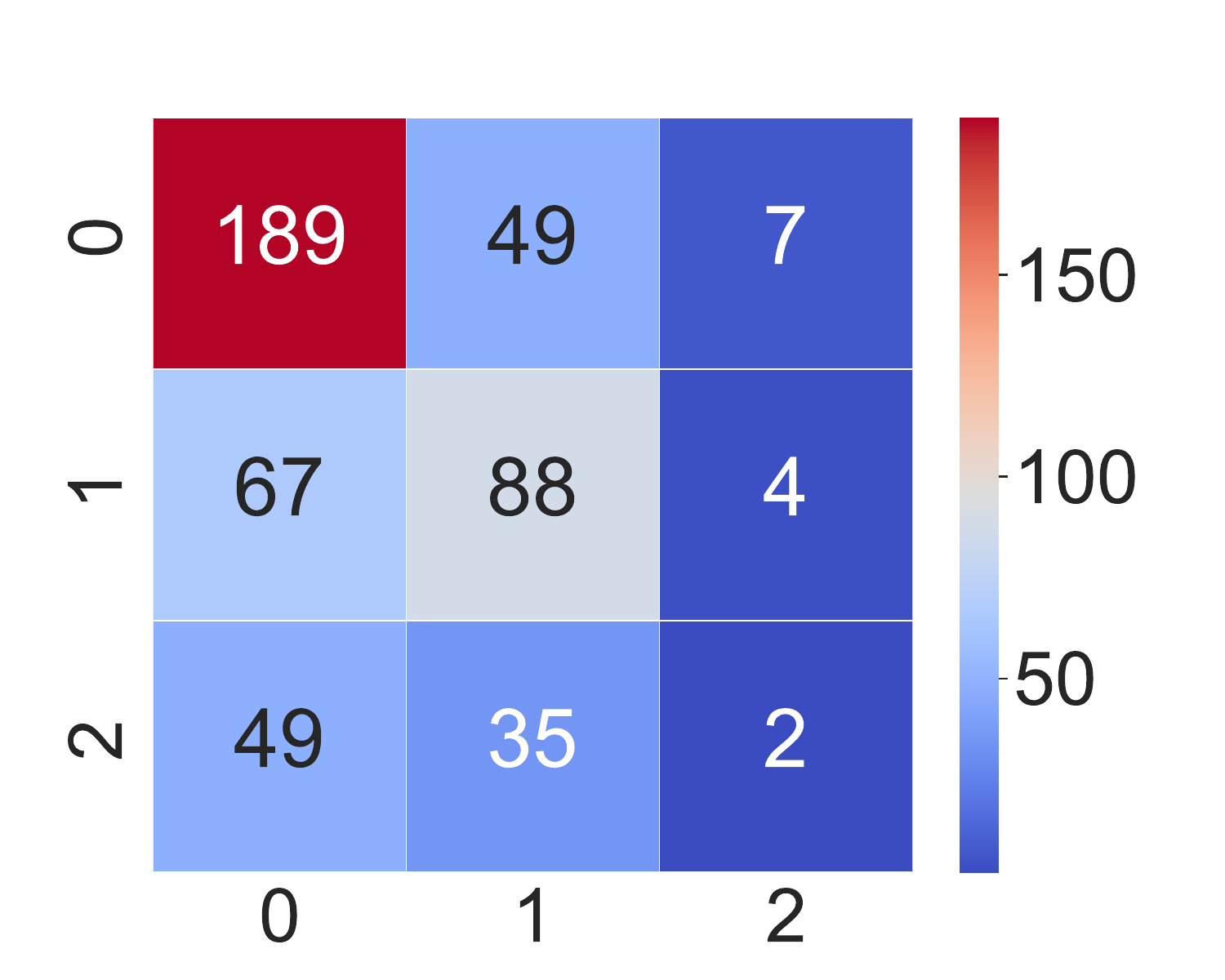}
	\caption{BiLSTM}
\end{subfigure}
\begin{subfigure}[b]{0.44\textwidth}
	\includegraphics[width=\textwidth]{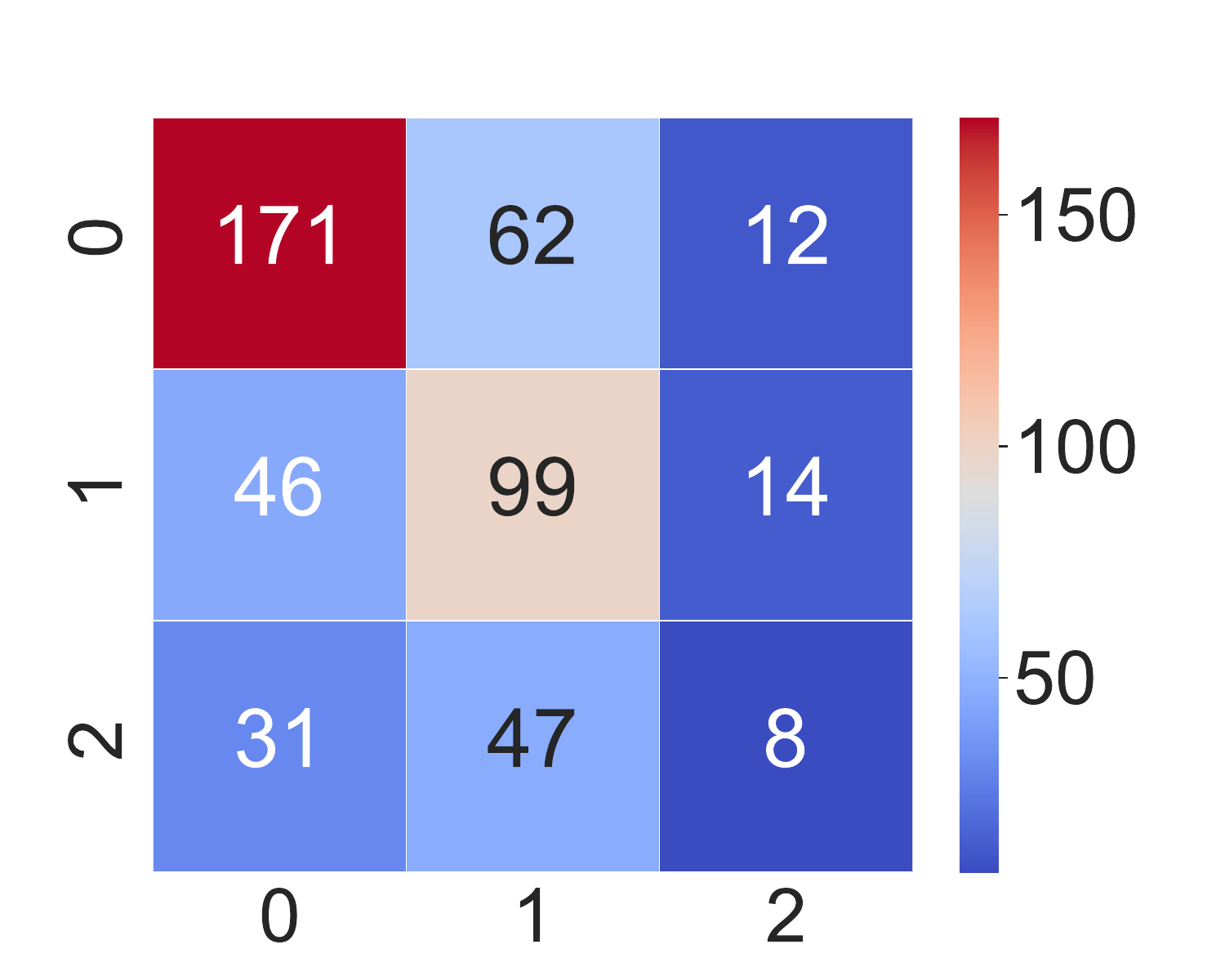}
	\caption{RN}
\end{subfigure}
\label{fig:cm}
\end{figure}

This section conducts error analysis, taking UMD dataset as an example. As mentioned before in the last section, most methods suffer from poor performance in predicting low-risk posts. Fig.~\ref{fig:cm} shows the heat maps of the confusion matrix of BiLSTM and our RN-based model, where axis 0, 1, and 2 represent labels of -1, 1, and 0. These two methods tend to predict more instances as none or high risk. Furthermore, our proposed method has a slightly better result than its counterpart. 
We also notice that our proposed model can achieve a higher accuracy of 59.18\% on the UMD dataset. However, it fails in predicting low-risk suicidal ideation, with a similar performance to other baselines. 
 
\paragraph{Limitations and Future~Work.}
We use sentiment lexicons and a topic model to exact sentiment- and topic-related risk indicators for relational text encoding. This preprocessing procedure can cause error propagation. Sentiment varies in different social communities. Using existing lexicons in popular communities may have a limitation. In future work, we will consider building lexicons from mental health-related communities. 
Machine learning-based based automatic detection systems cannot be treated as professional medical diagnoses.
However, they can empower professional practitioners and help them identify potential victims from many social posts.
The reasons for mental disorders and suicidal ideation are complex. 
Our future work will explore mental and suicidal factors and more effective relational reasoning to boost predictive performance. 

\section{Conclusion}
\label{sec:conclusion}
Text classification on mental disorders might not be treated as a medical diagnosis of professional practitioners. It can act as a computer-aided system to automatically provide early warnings of online social users at risk and notify social workers to provide early intervention.
It can also help the social workers and volunteers to identify the type of mental disorders, relieve online users' mental health issues through conversations, and suggest proper consultations or treatments.
This paper attempts to encode text by integrating suicidal ideation with sentimental indicators and life event-related topical indicators and proposes RNs with an attention mechanism for relational encoding. Experiments show the effectiveness of our proposed model. We argue that it is a significant step to combine canonical feature extraction with RNs for reasoning. 

\section*{Acknowledgments}
The authors would like to thank Philip Resnik for providing the UMD Reddit Suicidality Dataset and Mark Dredze for providing the dataset in the CLPsych 2015 shared task. This research is supported by the Australian Research Council (ARC) Linkage Project (LP150100671).

\bibliographystyle{unsrt}
\bibliography{sid-mh-rn}

\end{document}